\documentclass[sn-standardnature,iicol]{sn-jnl}

\jyear{2023}%

\usepackage{adjustbox}
\usepackage{chngcntr}
\usepackage{multirow}
\usepackage{makecell}
\usepackage{xcolor}         
\usepackage{color, colortbl}
\usepackage{float}
\usepackage{bbding}
\usepackage{pifont}
\usepackage{wasysym}
\usepackage{amssymb}

\theoremstyle{thmstyleone}%
\theoremstyle{thmstyletwo}%

\theoremstyle{thmstylethree}%

\begin{document}

\title[Article Title]{Mimic before Reconstruct: Enhancing Masked Autoencoders with Feature Mimicking}


\author[1]{\fnm{Peng} \sur{Gao}}\email{gaopeng@pjlab.org.cn}

\author[2]{\fnm{Renrui} \sur{Zhang}}\email{zhangrenrui@pjlab.org.cn}

\author[2]{\fnm{Rongyao} \sur{Fang}}\email{rongyaofang@link.cuhk.edu.hk}

\author[2]{\fnm{Ziyi} \sur{Lin}}\email{zylin@link.cuhk.edu.hk}

\author[1]{\fnm{Hongyang} \sur{Li}}\email{lihongyang@pjlab.org.cn}

\author[2]{\\\fnm{Hongsheng} \sur{Li}}\email{hsli@ee.cuhk.edu.hk}

\author[1]{\fnm{Qiao} \sur{Yu}}\email{yu.qiao@siat.ac.cn}

\affil[1]{\orgname{Shanghai AI Laboratory}, \orgaddress{\city{Shanghai}, \country{China}}}

\affil[2]{\orgdiv{Department of Electronic Engineering}, \orgname{CUHK}, \orgaddress{\state{Hong Kong SAR.}, \country{China}}}



\abstract{Masked Autoencoders (MAE) have been popular paradigms for large-scale vision representation pre-training. However, MAE solely reconstructs the low-level RGB signals after the decoder and lacks supervision upon high-level semantics for the encoder, thus suffering from sub-optimal learned representations and long pre-training epochs. To alleviate this, previous methods simply replace the pixel reconstruction targets of 75\% masked tokens by encoded features from pre-trained image-image (DINO) or image-language (CLIP) contrastive learning. Different from those efforts, we propose to \textbf{M}imic before \textbf{R}econstruct for Masked Autoencoders, named as \textbf{MR-MAE}, which jointly learns high-level and low-level representations without interference during pre-training. For high-level semantics, MR-MAE employs a mimic loss over 25\% visible tokens from the encoder to capture the pre-trained patterns encoded in CLIP and DINO. For low-level structures, we inherit the reconstruction loss in MAE to predict RGB pixel values for 75\% masked tokens after the decoder. As MR-MAE applies high-level and low-level targets respectively at different partitions, the learning conflicts between them can be naturally overcome and contribute to superior visual representations for various downstream tasks. On ImageNet-1K, the MR-MAE base pre-trained for only 400 epochs achieves 85.8\% top-1 accuracy after fine-tuning, surpassing the 1600-epoch MAE base by +2.2\% and the previous state-of-the-art BEiT V2 base by +0.3\%. Code and pre-trained models will be released at \url{https://github.com/Alpha-VL/ConvMAE}.}




\maketitle

\section{Introduction}\label{intro}

Masked Language Modeling (MLM)~\cite{devlin2018bert,brown2020language,radford2019language} has revolutionized natural language understanding via the large-scale pre-training. Motivated by this, Masked Autoencoders (MAE)~\cite{he2022masked} explore how to adopt MLM paradigm into vision representation learning with a vision transformer~\cite{dosovitskiy2020image} of asymmetric encoder-decoder architectures. MAE only encodes 25\% visible image tokens and reconstructs the RGB pixels values of other 75\% masked tokens. The representations learned through MAE have shown promising performances on various downstream vision tasks, which surpass the contrastive learning paradigms~\cite{radford2021learning,caron2021emerging,he2020momentum}. 

Although MAE is rising to be the dominant approaches for vision representation learning, it still suffers from the following disadvantages compared with its MLM counterparts. Firstly, the success of MLM pre-training~\cite{devlin2018bert} benefits from reconstructing the human-abstracted word tokens with rich semantics. It poses a non-trivial pre-text task that guides the transformer to learn informative representations for language understanding. Different from the high-level supervisions in language modeling, the low-level RGB signals of MAE~\cite{he2022masked} is too primitive and redundant, which fail to unleash the full understanding capacity of masked autoencoding on downstream vision tasks. Secondly, MAE~\cite{he2022masked} employs an asymmetric architecture with a heavy encoder and a light decoder, where the encoder is preserved after pre-training for downstream transfer learning. However, MAE only applies the pre-training supervision upon decoder's outputs, which are insufficient to guide the encoder and slows down the convergence speed of the pre-training stage.

To build more effective reconstruction targets, existing methods~\cite{wei2022masked,baevski2022data2vec,wei2022mvp,peng2022beit,hou2022milan} explore off-the-shelf pre-trained DINO~\cite{caron2021emerging}, CLIP~\cite{radford2021learning}, or online momentum features~\cite{he2020momentum} as the high-level supervisions. However, considering the interference caused by simultaneous reconstruction of different targets, previous methods simply replace the original RGB pixel targets by the high-level features and only use them to supervise the decoder's outputs.

\begin{figure*}[t]
\centering
\includegraphics[width=0.8\linewidth]{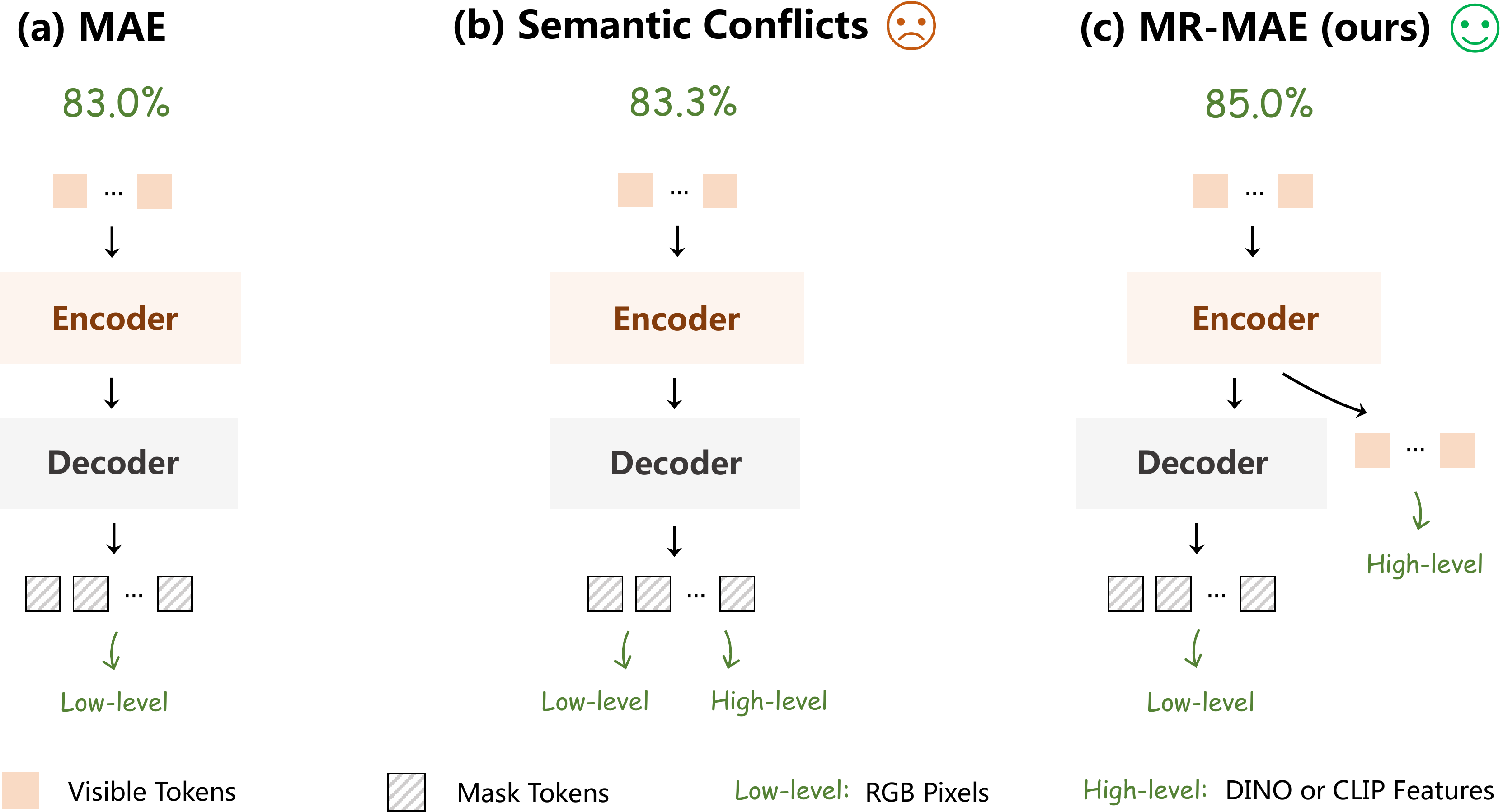}
\vspace{0.1cm}
\caption{\textbf{Pre-training with MR-MAE.} (a) The original MAE only reconstructs low-level RGB pixels for masked tokens. (b) Applying both low-level and high-level supervisions to the decoder outputs causes semantic conflicts. (c) Our MR-MAE applies low-level and high-level supervisions respectively to different image tokens and network layers. The top-1 accuracy by fine-tuning on ImageNet-1K~\cite{russakovsky2015imagenet} can be improved from 83.0\% to 85.5\%.}
\label{fig: main_img}
\end{figure*}

Different from previous approaches~\cite{wei2022masked,baevski2022data2vec,wei2022mvp,peng2022beit,hou2022milan} that only apply high-level supervisions at the decoder, we aim to take advantages of both high-level semantics and low-level textures, and benefit the encoder's pre-training by learning with the two targets. To overcome the conflicts between the two types of semantics, we introduce a new framework, named Mimic-before-Reconstruct Masked Autoencoders (MR-MAE). The original MAE randomly samples 25\% visible tokens and processes them by the encoder. Then, the encoded tokens mixed with position embeddings are fed into the light-weight decoder for predicting pixel values of the 75\% masked tokens. Our proposed MR-MAE augments the original MAE with a simple yet effective mimic loss~\cite{hinton2015distilling}, which is applied to only the visible tokens directly after the encoder. The mimic loss minimizes the L2 distance between the MAE encoder's outputs and the high-level features generated from off-the-shelf pre-trained image-language (CLIP)~\cite{radford2021learning} or image-image (DINO)~\cite{caron2021emerging}. Unlike the insufficiently supervised encoder in MAE, such mimic loss can provide effective and direct guidance on the encoder. As our mimic loss and the reconstruction loss are applied for different groups of tokens (25\% visible vs 75\% masked) and different network layers (encoder vs decoder's outputs), our MR-MAE well solves the supervision conflicts between the low-level and high-level learning targets.

Compared with the original MAE base model~\cite{he2022masked} (83.8\%) that aims to reconstruct low-level RGB pixels, our MR-MAE base model with a CLIP teacher not only enhances the ImageNet-1K fine-tuning accuracy to 85.8\% (+2.0\%), but also shortens the pre-training epochs from 1600 to 200. Notably, MR-MAE base and huge-392 models surpass the ImageNet-1K~\cite{russakovsky2015imagenet} fine-tuning accuracy of CLIP (84.2\%) by +1.4\% and +4.1\%, respectively. This indicates that our MR-MAE learns even better representations than the high-level teacher network, other than being upper-bounded, demonstrating the benefit to jointly learn low-level and high-level targets.

\section{Related Work}
\subsection{Contrastive Learning}
Contrastive learning~\cite{chen2020simple,wu2018unsupervised,caron2021emerging,radford2021learning} has achieved great successes on learning effective visual representations by extracting invariances from augmented views of a signal source. DINO~\cite{caron2021emerging} and CLIP~\cite{radford2021learning} are two canonical approaches among contrastive learning paradigms. DINO~\cite{caron2021emerging} observed strong objectness emerges from ViT pre-trained by image-image contrastive learning. On the other hand, CLIP~\cite{radford2021learning} demonstrated amazing zero-shot ability through image-text pair contrastive learning. Although DINO and CLIP exhibits strong objectness cues and open-world recognition ability, the fine-tuning performance on downstream tasks are inferior to representations learned through MAE~\cite{he2022masked} manner. Our MR-MAE borrows the high-level semantics extracted from off-the-shelf DINO or CLIP to supervise the features of visible tokens in MAE. Thanks to the guidance of teacher networks, MR-MAE can significantly improve the representations of MAE and shorten the training epochs. 

\subsection{Masked Image Modeling}
Pre-training on large-scale unsupervised corpus with Masked Language Modeling (MLM)~\cite{devlin2018bert} have shown superior performance on natural language understanding and generation. Motivated by MLM, BEiT~\cite{bao2021beit} explored Masked Image Modeling (MIM) on vision transformers by reconstructing the vision dictionary extracted with DALL-E~\cite{ramesh2021zero,ramesh2022hierarchical}. MAE~\cite{he2022_mae} further proposed an asymmetric encoder and decoder for scaling up MIM to huge models. Besides, it demonstrated a simple pixel reconstruction loss can learn good visual representations. Due to the simplicity and computational efficiency, MAE is raising to a popular generative pre-training paradigm. As MAE reconstructs low-level signals with an isotropic vision transformer architecture, researchers improve MAE by exploring high-level signals and hierarchical architectures. MaskFeat~\cite{wei2022masked}, data2vec~\cite{baevski2022data2vec}, MVP~\cite{wei2022mvp} and MILAN~\cite{hou2022milan} revealed various high-level signals, such as pre-trained DINO~\cite{caron2021emerging}, HOG features~\cite{dalal2005histograms}, momentum features~\cite{he2020momentum} and multi-modality features~\cite{radford2021learning}, which are more effective than reconstructing low-level signals. Different from those approaches that explore high-level features as new reconstruction targets of masked regions, MR-MAE utilizes high-level features for regularizing the representations of visible tokens produced by MAE encoder. Thus, our MR-MAE can take advantages of both low-level and high-level information. FD~\cite{wei2022contrastive} proposed to improve pre-trained contrastive representations through feature distillation. Compared with FD to feed all tokens into the encoder, the encoder of MR-MAE only processes partially visible tokens (e.g., 25\%) which leads to a significantly decrease of GPU memory. DMAE \cite{bai2022masked} proposed to jointly optimize the reconstruction loss and align the features with pre-trained MAE teacher. As the MAE teacher is still pre-trained by reconstructing low-level signals, the representations of DAME still lack high-level semantics. Different from DMAE, MR-MAE guides the feature distillation with contrastively pre-trained features which are complementary with low-level signals. MCMAE~\cite{gao2022convmae}, UM-MAE~\cite{li2022uniform}, MixMIM~\cite{liu2022mixmim} and GreenMIM~\cite{huang2022green} explore efficient and effective MIM frameworks with hierarchical vision transformers~\cite{liu2021swin,gao2021container,li2022uniformer,xiao2021early}. Our MR-MAE also leverages the masked convolution stages in MCMAE to hierarchically encode visual representations.

\section{Method}
\subsection{Revisiting MAE}
Masked Autoencoders (MAE)~\cite{he2022masked} employ an asymmetric encoder-decoder design for computationally efficient masked image modeling. Given an input image, MAE first divides it into patches of size $p \times p$, and randomly masks 75\% of them. We denote the masked and visible patches respectively as $I_m \in \mathbb{R}^{l_m \times p^2}$ and $I_v \in \mathbb{R}^{l_v \times p^2}$, where $l_m$ and $l_v$ denote the numbers of masked and visible tokens. Then, the 25\% visible patches are tokenized and fed into a transformer encoder to produce the $C$-dimensional intermediate representation $E_v \in \mathbb{R}^{l_v \times C}$. As shown in Figure~\ref{fig: main_img} (1), MAE employs a light-weight transformer decoder to predict $D_m \in \mathbb{R}^{l_m \times p^2}$ to reconstruct RGB values of the masked tokens. An L2 reconstruction loss $\mathcal{L_{R}}$ between $D_m$ and $I_m$ is used:
\begin{equation}
    \mathcal{L_{R}} = \frac{1}{l_m} \|D_m - I_m\|_2^2.
\label{eq1}
\end{equation}

Despite its promising transfer capacity, MAE requires costly 1600 epochs to be fully pre-trained, which is partially due to the missing guidance to the intermediate representations of the encoder. Furthermore, by visualizing the attention map of [CLS] tokens in MAE's encoder as shown in Figure~\ref{fig: attn_vis}, we observe that MAE focuses more on some detailed texture patterns than the centric objects, since merely low-level RGB values $I_m$ serve as the reconstruction targets. Therefore, we argue that the low-level supervision at the decoder's outputs not only slows down the pre-training convergence of MAE, but also limits its representations to capture high-level semantics.


\begin{figure*}[t]
\centering
\includegraphics[width=0.9\linewidth]{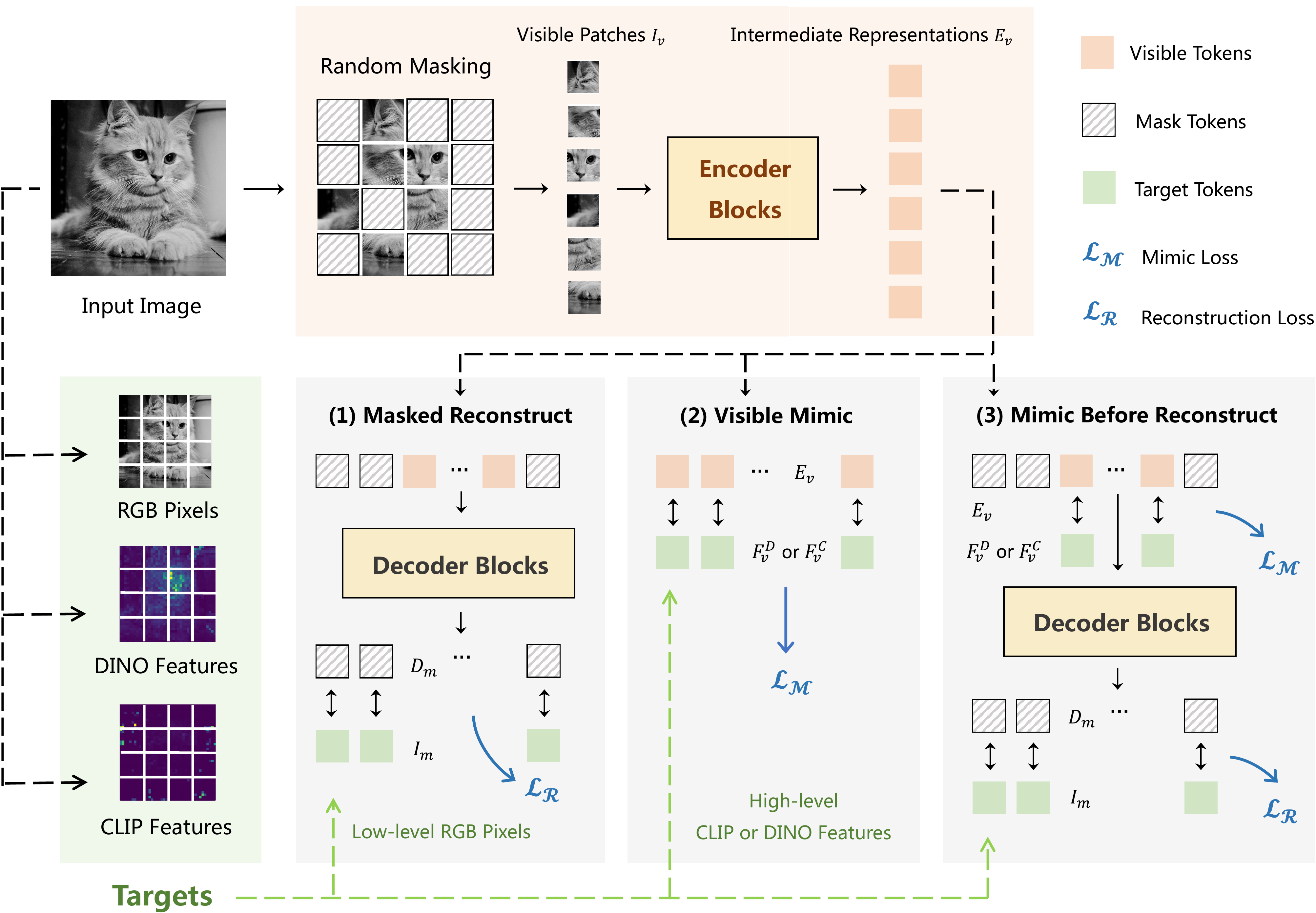}
\vspace{0.1cm}
\caption{\textbf{Architecture of MR-MAE.} During MAE pre-training, we set both high-level and low-level learning targets respectively for different image tokens and network layers: mimic loss for 25\% visible tokens of the encoder, and reconstruction loss for 75\% masked tokens of the decoder.}
\label{fig: main_img}
\end{figure*}

\subsection{Mimic before Reconstruct}

To address the above issues, we propose to \textbf{M}imic before \textbf{R}econstruct for Masked Autoencoders, termed as MR-MAE, which is a simple and effective strategy to enhance MAE~\cite{he2022masked} by regularizing the intermediate representations with pre-trained off-the-shelf feature encoders. The overall pipeline of MR-MAE is illustrated in Figure~\ref{fig: main_img}. Following MAE, MR-MAE also inputs the visible 25\% tokens into the transformer encoder to encode the intermediate representation $E_v$. Different from only supervising the low-level reconstruction after the decoder, we propose to guide the intermediate $E_v$ from the encoder with high-level features produced by DINO or CLIP, which contain rich high-level semantics, as shown in Figure~\ref{fig: main_img} (2). We first extract the DINO or CLIP features by feeding the input image into their transformer-based visual encoders, denoted as $F_v^{D}, F_v^{C} \in \mathbb{R}^{l_v \times C}$. By appending a feature mimic head on top of the encoder, we transform the visible representations $E_v$ via a linear projection layer to mimic $F_v^{D}$ or $F_v^{C}$. 
The L2 mimic loss of MR-MAE is defined as:
\begin{equation}
    \mathcal{L_{M}} = \frac{1}{l_v}\|\operatorname{L}(E_v) - F_v\|^2_2,
\end{equation}
where $F_v$ denotes either DINO's $F_v^{D}$ or CLIP's $F_v^{C}$ while L denotes mimic head.

To incorporate both low-level and high-level information, we also apply a light-weight decoder in MR-MAE to reconstruct the 75\% masked RGB pixels, as shown in Figure~\ref{fig: main_img} (3). We adopt the L2 reconstruction loss $\mathcal{L_{R}}$ in Eq.~\ref{eq1} between $D_m$ and $I_m$. As the feature mimic loss $\mathcal{L_{M}}$ for visible tokens and the reconstruction loss $\mathcal{L_{R}}$ for masked tokens aim at encoding different aspects of the input image, i.e., high-level semantics and low-level textures, they can complement each other to learn more discriminative representations. In addition, MR-MAE avoids the conflict of learning between low-level and high-level targets by applying supervisions upon different groups of tokens (25\% visible vs 75\% masked) and different network layers (encoder vs decoder's outputs). 
With the newly introduced high-level feature mimic loss, our proposed MR-MAE significantly improves the downstream performance of MAE and  shortens its pre-training epochs.

\subsection{Bag-of-tricks for MR-MAE}
\label{tricks}
To further unleash the learning potential, we borrow some tricks from previous approaches and integrate them into MR-MAE to enhance our learned representations.

\paragraph{Focused Mimicking.}
MAE adopts a random masking strategy for visible token selection, which is a natural choice for low-level signal reconstruction without additional guidance. As the [CLS] token in off-the-shelf pre-trained models can clearly delineate regions of importance~\cite{caron2021emerging} via its attention map, we select the most salient tokens in teacher network's attention maps for visible feature mimicking. In this way, MR-MAE can better capture informative high-level semantics encoded in the teacher network, rather than the non-salient low-level ones. Similar strategies were previously discussed in MST~\cite{li2021mst}, ADIOS~\cite{shi2022adversarial}, AttnMASK~\cite{kakogeorgiou2022hide}, and MILAN~\cite{hou2022milan}.

\begin{table*}[t!]
\centering
\caption{Image classification by fine-tuning on ImageNet-1K~\cite{russakovsky2015imagenet}. `Ratio' denotes the visible ratio of image tokens fed into the encoder. `P-Epochs' and `FT' denote pre-training epochs and the top-1 accuracy by fine-tuning.}
\vspace{0.2cm}
\label{tab:imagenet}
\begin{adjustbox}{width=0.85\linewidth}
\begin{tabular}{l|c|cccc|c}
	\toprule
Methods & Backbone & Params. (M) & Supervision & Ratio  & P-Epochs & FT (\%) \\
\midrule
BEiT~\cite{bao2021beit} & ViT-B & 88 & DALLE & 100\%  & 300 & 83.0 \\
        MAE~\cite{he2022masked} & ViT-B & 88 & RGB & 25\%  & 1600 & 83.6 \\
        CAE~\cite{chen2022context} & ViT-B & 88 & RGB & 25\% & 800 & 83.6 \\
        MaskFeat~\cite{wei2022masked} & ViT-B & 88 & HOG & 100\%  & 300 & 83.6 \\
        SimMIM~\cite{xie2022simmim} & Swin-B & 88 & RGB & 100\%  & 800 & 84.0 \\
        DMAE~\cite{bai2022masked} & ViT-B & 88 & MAE & 25\% & 100 & 84.0 \\
        data2vec~\cite{baevski2022data2vec} & ViT-B & 88 & Momentum & 100\%  & 800 & 84.2 \\
        MVP~\cite{wei2022mvp} & ViT-B & 88 & CLIP & 100\%  & 300 & 84.4 \\
        MCMAE~\cite{gao2022convmae} & CViT-B & 88 &   RGB & 25\%  & 1600 & 85.0 \\
        MixMIM~\cite{liu2022mixmim} & MixMIM-B & 88 & RGB & 100\% & 600 & 85.1 \\
        CMAE~\cite{huang2022contrastive} & CViT-B & 88 &   RGB & 25\%  & 1600 & 85.3 \\
        MILAN~\cite{hou2022milan} & ViT-B & 88 & CLIP & 25\% & 400 & 85.4 \\
        BEiT V2~\cite{peng2022beit} & ViT-B & 88 & CLIP & 100\% & 1600 & 85.5 \\
        \midrule
        MR-MAE & CViT-B & 88 & CLIP & 25\%  & 400 & 85.8 \\
        \bottomrule
\end{tabular}
\end{adjustbox}
\end{table*}

\paragraph{Multi-layer Fusion.}
The original MAE only feeds the output tokens from the encoder's last layer into the decoder for masked pixel reconstruction. As different layers of the encoder might depict different abstraction levels of an image, we fuse the visible tokens from multiple intermediate layers of the encoder by element-wisely addition, and then utilize the fused ones for high-level feature mimicking and low-level pixel reconstruction. By this, the supervision from feature mimicking can be directly applied to multiple layers of the encoder, leading to the improved visual representations. Similar results have been demonstrated in BERT~\cite{shi2022revisiting}, contrastive learning~\cite{wang2022repre}, and hierarchical MIM~\cite{gao2022convmae}.


\paragraph{Masked Convolution Stages.}
Exploring multi-scale visual information has achieved great successes on computer vision tasks as objects exist in various scales. Following  MCMAE~\cite{gao2022convmae}, we append extra masked convolution stages before the transformer blocks~\cite{gao2022convmae,xiao2021early,gao2021container,guo2022cmt} to efficiently capture high-resolution details, and apply multi-scale block-wise masking~\cite{gao2022convmae} to prevent information leakage for pixel reconstruction. Such multi-scale encoding can learn hierarchical representations and achieve significant improvements on downstream tasks.

\section{Experiments}

For image classification, we pre-train MR-MAE on ImageNet-1K~\cite{russakovsky2015imagenet}, and compare with state-of-the-art Masked Image Modeling (MIM) methods by fine-tuning for top-1 accuracy. To further evaluate MR-MAE on high-resolution images, we fine-tune our pre-trained model on COCO~\cite{lin2014microsoft} with Mask-RCNN~\cite{he2017mask} framework, and report $AP^{box}$ and $AP^{mask}$ results. 
Then, we conduct extensive ablation studies over each component of MR-MAE to validate their effectiveness.

\begin{table*}[t!]
\centering
\caption{Object Detection by fine-tuning on COCO~\cite{lin2014microsoft} based on the Mask-RCNN~\cite{he2017mask} framework. `F-epochs' denotes the epochs for fine-tuning.}
\label{tab:coco}
\begin{adjustbox}{width=0.83\linewidth}
\begin{tabular}{l|cc|cc|cc}
	\toprule
 Methods  &   P-Epochs & F-Epochs & $AP^{\rm box}$ & $AP^{\rm mask}$ &  Params. (M) & FLOPs (T)\\
        \midrule
        ViTDet~\cite{li2022exploring} &   1600 & 100 & 51.2 & 45.5  & 111 & 0.8\\
        CMAE~\cite{huang2022contrastive} &   1600 & 25 & 52.9 & 47.0 & 104 & 0.9\\
        MCMAE~\cite{gao2022convmae} &   1600 & 25 & 53.2 & 47.1 & 104 & 0.9\\
        \midrule
        MR-MAE &  400 & 25 & 53.4 & 46.9 & 104 & 0.9\\
        \bottomrule
\end{tabular}
\end{adjustbox}
\end{table*}

\begin{table*}[t]
    \centering
    \vspace{0.2cm}
    \caption{Ablation study for `mimic before reconstruct' and the bag-of-tricks for MR-MAE.}
    \begin{adjustbox}{width=0.83\linewidth}
    \begin{tabular}{c|ccccc|c|cc}
    \toprule
        \multirow{2}*{P-Epochs} &Low & High & Focused & Multi-layer & Masked   &ImageNet-1K & \multicolumn{2}{c}{COCO}\\
        & Level & Level & Mimic & Fusion & Conv.  &FT &AP$^{box}$ &AP$^{mask}$  \\
        \midrule
         \multirow{7}*{200} &$\mathcal{L_{R}}$ &  &  &  &  &  83.0 & N/A & N/A\\
         &$\mathcal{L_{R}}$ & $\mathcal{L_{M}}$ &  &  &  &  84.7 & N/A & N/A\\
         &$\mathcal{L_{R}}$ & $\mathcal{L_{M}}$  & \ding{51} &  &  & 84.9 & N/A & N/A\\
         &$\mathcal{L_{R}}$ & $\mathcal{L_{M}}$  & \ding{51} & \ding{51} &  & 85.0 &  51.6 & 45.5\\
         &$\mathcal{L_{R}}$ & $\mathcal{L_{R}}$  & \ding{51} & \ding{51} &  & 83.3 &  50.3 & 44.9\\
         & & $\mathcal{L_{M}}$ & \ding{51} & \ding{51} &  & 84.9 & 50.9 & 44.8\\
         &$\mathcal{L_{R}}$ & $\mathcal{L_{M}}$ & \ding{51} & \ding{51} & \ding{51} & 85.5 &  53.0 & 46.5\\
        \bottomrule
    \end{tabular}
    \label{tab:many_ablation}
    \end{adjustbox}
\end{table*}

\subsection{ImageNet-1K Pre-training and Fine-tuning}
\paragraph{Experiment Setups.}
For MR-MAE, we follow the protocol of pre-training and fine-tuning on ImageNet-1K as previous approaches. Specifically, our MR-MAE base is pre-trained for 400 epochs with batch size 1,024 and weight decay 0.05. We adopt the AdamW~\cite{loshchilov2018fixing} optimizer and the cosine learning rate scheduler with an maximum learning rate $1.5 \times 10^{-4}$ and 80-epoch warming up. We utilize the mask ratio 25\% and 8 decoder blocks following the practices in MAE~\cite{he2022masked}. The pre-training of MR-MAE jointly optimizes the reconstruction loss and mimic loss, whose weights are 0.5 and 0.5. By default, we choose ViT-B/16 pre-trained by CLIP~\cite{radford2021_clip} as the high-level teacher. After the self-supervised pre-training, we transfer the pre-trained encoder as an initialization for fine-tuning on ImageNet-1K and report the top-1 accuracy on the validation set. The fine-tuning takes 100 epochs with 5-epoch warming up. We adopt the same batch size, optimizer, and weight decay as pre-training. The initial learning rate, layer-wise learning rate decay, and drop path rate are set to be $3\times 10^{-4}$, 0.6 and 0.2, respectively. 

\paragraph{Results on ImageNet-1K Fine-tuning.}
We compare our MR-MAE base model with previous state-of-the-art approaches of the similar model size on Table~\ref{tab:imagenet}. BeiT~\cite{bao2021beit}, MAE~\cite{he2022masked}, CAE~\cite{chen2022context} have validated Masked Image Modeling (MIM) paradigm to be effective approaches for pre-training vision transformers. Due to their reconstruction of low-level pixels and the adoption of isotropic architectures, our MR-MAE can surpass the performance of those approaches by large margins (85.8\% vs 83.0/83.6/83.6/84.0\%). SimMIM~\cite{xie2022simmim}, MCMAE~\cite{gao2022convmae} and MixMIM~\cite{liu2022mixmim} introduce multi-scale features into MIM, resulting in improved fine-tuning accuracy compared with the isotropic architectures. As previous multi-scale approaches still reconstruct low-level signals, our MR-MAE can surpass their fine-tuning accuracy (85.8\% vs 84.0/85.0/85.1\%) with fewer pre-training epochs (400 vs 800/1600/600). 

Another line of researches focuses on directly replacing the reconstruction of low-level signals with high-level semantic targets, MaskFeat~\cite{wei2022masked}, data2vec~\cite{baevski2022data2vec}, MVP~\cite{wei2022mvp} and MILAN~\cite{hou2022milan} demonstrate promising results by integrating DINO~\cite{zhang2022dino}, momentum features~\cite{he2020momentum} and CLIP~\cite{radford2021_clip}. MILAN~\cite{hou2022milan} proposes a novel promoting decoder and semantic-aware masking to enhance the feature learning by reconstructing high-level features. BeiT V2~\cite{peng2022beit} replaces the original DALL-E tokenizers with high-level semantic tokenizers learned by self-encoding of CLIP features. Compared with advanced approaches for reconstructing high-level signals, such as MILAN and BeiT-V2, MR-MAE still achieves better performance (85.8\% vs 85.4/85.5\%), since we jointly learn low-level and high-level targets with multi-scale architectures. CMAE~\cite{huang2022contrastive} learns representations through joint optimization of contrastive loss and reconstruction loss. Different from CMAE, MR-MAE utilizes a teacher model pre-trained from large-scale image-text contrastive learning, which contains more abundant semantic knowledge. MR-MAE improves the top-1 accuracy of CMAE from 85.3\% to 85.8\% and shortens the pre-training epochs from 1600 to 400. DMAE~\cite{bai2022masked} adopts a similar approach as MR-MAE, which mimics features generated from the pre-trained teacher and reconstructs the low-level pixels. However, since the teacher of DMAE is still pre-trained with low-level pixel targets, the fine-tuning accuracy of DMAE is inferior to MR-MAE (84.0\% vs 85.8\%).

\begin{table*}[t!]
\begin{minipage}[h!]{0.48\linewidth}
\centering
\small
\caption{Ablation study for high-level pre-training targets on the ImageNet-1K fine-tuning accuracy.}
\begin{adjustbox}{width=0.8\linewidth}
\label{tab:target_ablate}
\begin{tabular}{c|cc}
	\toprule
High-level Target & Params (M) & FT \\
          \midrule
         DINO & 88  &  84.0  \\
         CLIP & 88  &  85.0  \\
         CLIP/DINO (Joint) & 88  &  83.8  \\
         CLIP/DINO (Sep.)  & 176 &  85.5  \\
        \bottomrule
\end{tabular}
\end{adjustbox}
\end{minipage}\quad
\begin{minipage}[h!]{0.49\linewidth}
\centering
\small
\caption{Ablation study for the influence of pre-training epochs on ImageNet-1K and COCO object detection.}
\begin{adjustbox}{width=0.9\linewidth}
\label{tab:epochs}
\begin{tabular}{c|c|cc}
	\toprule
 \multirow{2}*{P-Epochs} &ImageNet-1K & \multicolumn{2}{c}{COCO} \\
&FT  &$AP^{box}$ &$AP^{mask}$ \\ \midrule
         200 & 85.5  & 53.0 & 46.5  \\
         400 & 85.8  & 53.4 & 46.9  \\
         800 & 85.8  & 53.5 & 47.0  \\
        \bottomrule
\end{tabular}
\end{adjustbox}
\end{minipage}
\end{table*}

\subsection{Object Detection}
\paragraph{Experiment Setups.}
We evaluate the downstream transfer capacity of MR-MAE on the widely adopted COCO dataset~\cite{lin2014microsoft}. We apply the pre-trained encoder of MR-MAE as initialization of backbone for Mask-RCNN. Following ViTDet~\cite{li2021benchmarking,li2022exploring}, we simply expand the features for multiple scales as an alternative of feature pyramid network (FPN)~\cite{lin2017feature}. The resolution of the input image, learning rate, and layer decay are set as $1,024 \times 1,024$, $2\times 10^{-4}$ and 0.8, respectively. The model is fine-tuned for 25 epochs with batch size 16.


\paragraph{Results on COCO Fine-tuning.}
In Table~\ref{tab:coco}, we use our proposed MR-MAE as the pre-trained backbone for Mask-RCNN~\cite{he2017mask}. MR-MAE attains 53.4\% $AP^{box}$ and 46.9\% $AP^{mask}$ by fine-tuning 25 epochs on the COCO train2017 split. Compared with the baseline ViTDet~\cite{li2022exploring}, which adopts the encoder of MAE pre-trained for 1600 epochs, MR-MAE can improve $AP^{box}$ and $AP^{mask}$ by +2.2\% and +1.4\%. Besides, we shorten the pre-training epochs from 1600 to 400 and the fine-tuning epochs from 100 to 25. Compared with multi-scale backbones, such as CMAE~\cite{huang2022contrastive} and MCMAE~\cite{gao2022convmae}, MR-MAE achieves comparable $AP^{box}$ and $AP^{mask}$ with a much shorter pre-training epochs (1600 vs 400 epochs).

\subsection{Ablation Studies}
To validate each component of MR-MAE, we conduct the following ablation studies. 

\paragraph{Mimic Before Reconstruct.}
As shown in the first row of Table~\ref{tab:many_ablation}, the baseline MAE model with the low-level reconstruction loss achieves 83.0\% fine-tuning accuracy on ImageNet-1K with 200-epoch pre-training. By jointly learning with the mimic loss, the classification accuracy is boosted by +1.7\%. The comparison between the forth and sixth rows of Table~\ref{tab:many_ablation} indicates that the joint optimization of both low-level and high-level targets can achieve better performance than only mimicking high-level semantics, especially for $AP^{box}$ of object detection (+0.7\%). 
\begin{figure*}[!t]
\centering
\includegraphics[width=0.95\linewidth]{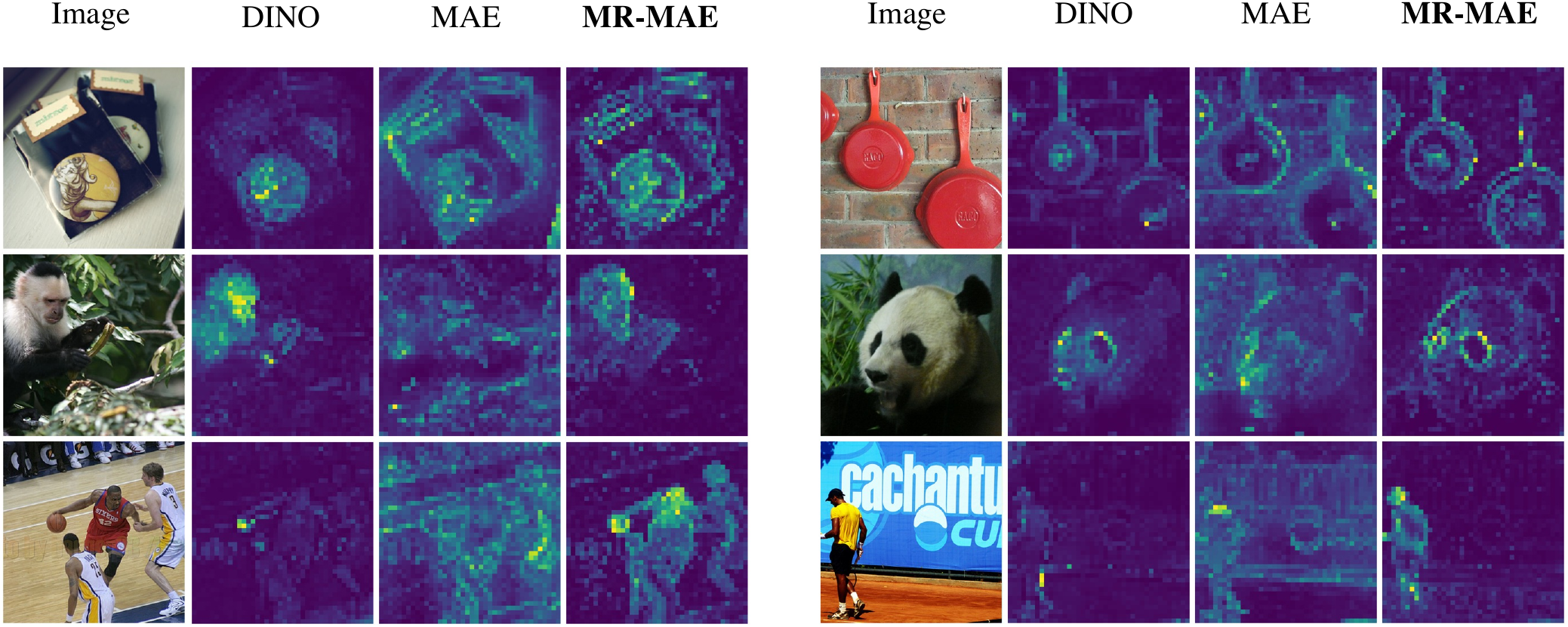}
\caption{Visualization of attention weights at the last self-attention layer in DINO \cite{zhang2022dino}, MAE \cite{he2022_mae}, and \textbf{MR-MAE (ours)}. MR-MAE can better capture salient feature representation compared to previous methods.}
\label{fig: attn_vis}
\end{figure*}

\begin{table*}[t!]
\centering
\vspace{0.1cm}
\caption{ImageNet-1K fine-tuning accuracy of different model scales.}
\label{tab:scale}
\begin{adjustbox}{width=0.85\linewidth}
\begin{tabular}{l|c|cccccc}
	\toprule
 Method &P-Epochs  & Small & Base & Large & Huge & Huge-393 & Huge-448\\
          \midrule
          MAE~\cite{he2022_mae} & 1600 & 79.5 & 83.6 & 85.9& 86.9 & - & 87.8 \\
          MCMAE~\cite{gao2022convmae} & 800 & 82.6 & 84.6 & 84.9 &86.2 & - & - \\
          MR-MAE & 200 & 83.6 & 85.5 & 86.8 &88.0 & 88.5 & -\\
        \bottomrule
\end{tabular}
\end{adjustbox}
\vspace{0.2cm}
\end{table*}

\paragraph{Bag-of-tricks.}
In Table~\ref{tab:many_ablation}, we also ablate each trick mentioned in Section~\ref{tricks}. Based on the 84.7\% fine-tuning accuracy with both $\mathcal{L_{R}}$ and $\mathcal{L_{M}}$, Focused mimicking leads to +0.2\% improvement due to the focus of salient tokens guided by attention maps of the teacher network. Multi-layer Fusion further improves the accuracy by +0.1\%. The introduction of Masked Convolution Stages increases the ImageNet-1K fine-tuning accuracy by +0.5\%. More importantly, it improves $AP^{box}$ and $AP^{mask}$ by +1.4\% and +1.0\%, respectively, demonstrating the significance of multi-scale architectures.

\paragraph{Conflicts between Low-level and High-level Targets.}
As low-level and high-level targets contain different visual semantics, their joint supervisions might conflict with each other. As shown in the forth and fifth rows of Table~\ref{tab:many_ablation}, joint reconstruction of low-level and high-level targets deteriorates ImageNet-1K fine-tuning accuracy by -1.7\%, $AP^{box}$ by -1.3\% and $AP^{mask}$ by -0.6\%. The results indicate our Mimic-before-Reconstruct framework is able to solve the conflicts between low-level and high-level targets by applying mimic and reconstruction losses upon different groups of tokens (visible vs masked) and different network layers (encoder vs decoder's outputs).  

\paragraph{Different High-level Targets.}
As image-image contrastive learning (DINO) and image-language contrastive learning (CLIP) encode different high-level semantics. We ablate the performance of MR-MAE base with different high-level semantics. As shown in Table~\ref{tab:target_ablate}, features generated by CLIP can surpass DINO by +1\%. This implies image-language contrastive learning provides stronger high-level semantics than image-image contrasive learning. The joint mimicking of multiple high-level signals is worse than independent mimicking. We hypothesize that the performance degradation is due to the gradient conflicts of predicting different high-level targets. To avoid the degradation introduced by the conflicts of reconstructing different high-level targets, we separately pre-train and fine-tune MR-MAE with different high-level targets then ensemble the two models. As shown in Table ~\ref{tab:target_ablate}, CLIP/DINO (Sep.) can surpass the CLIP/DINO (Joint) by +1.7\%, which validates the complementary representation learned with different targets. In the future, we will explore more efficient approaches to better incorporate multiple pre-trained high-level signals into a single student network.

\paragraph{Longer Pre-training Epochs.}
We ablate the influence of pre-training epochs on MR-MAE in Table~\ref{tab:epochs}. MR-MAE pre-trained for 200 epochs can achieve 85.5\% ImageNet-1K fine-tuning accuracy and 52.7\% $AP^{box}$ for COCO. MR-MAE pre-trained for 400 epochs can improve ImageNet-1K fine-tuning accuracy by +0.3\% and $AP^{box}$ by +0.7\%. Given longer pre-training epochs, such as 800 epochs, the performance saturates as shown in Table~\ref{tab:epochs}. This implies the introduction of high-level targets can make MIM approach converge much faster. The previous prolonged 1600 pre-training schedule can be shorted to 400 epochs under our Mimic-before-reconstruct framework. 

\paragraph{Scaling-up the Model.}
To test the scalability of our framework, we experiment with different models size of MR-MAE and reported the ImageNet-1K fine-tuning accuracy on Table~\ref{tab:scale}. Compared with the single-scale baseline MAE and the stronger multi-scale baseline MCMAE, our MR-MAE demonstrates significantly improved performance over all model sizes with much shortened pre-training epochs.

\paragraph{Feature Visualization.}
To provide intuitions on why high-level targets improve the representation, we visualize the attention map of [CLS] token of the last self-attention layer of different models. As shown in  Figure~\ref{fig: attn_vis}, the attention of MAE is biased towards texture patterns due to its aim of low-level pixel reconstruction, implying that MAE waste its capacity on low-level textures irrelevant for semantic understanding. On the other side, the attention of DINO's [CLS] token overemphasises on partial information of salient object. The attention of our MR-MAE can capture complete object information compared with DINO and MAE.

\section{Conclusion}
In this paper, we propose MR-MAE, a simple and effective framework for masked image modeling, which conducts feature mimicking before pixel reconstruction to incorporate high-level semantics into MAE. Specifically, for the 25\% visible tokens from the encoder, we apply a mimic loss upon them to learn the semantic information encoded by off-the-shelf pre-trained models. For the 75\% masked tokens after the decoder, we preserve the original reconstruction loss to model low-level texture patterns. By this, our MR-MAE does not only model both high-level and low-level information, but also well solves the semantic conflicts between the two types of targets, achieving superior performance for image classification and downstream detection.  

\paragraph{Limitation:}
Although MR-MAE effectively learns the high-level knowledge from CLIP or DINO, naive joint supervision of CLIP and DINO cannot achieve higher results (separate supervision first and model ensemble later can improve). Our future direction will focus on how to better guide MAE by high-level semantics from multiple teacher networks.

\bibliography{sn-article}


\end{document}